\pdfoutput=1
\documentclass[11pt, letterpaper, shortlabels]{berkeley}

\usepackage{hyperref}
\usepackage{color-edits}

\usepackage{algorithm2e}
\usepackage{algorithm}
\usepackage{algorithmicx}
\usepackage{algpseudocode}

\ifx\assumption\undefined
\newtheorem{assumption}{Assumption}
\fi

\usepackage[capitalize,noabbrev]{cleveref}
\makeatletter
\def\adl@drawiv#1#2#3{%
        \hskip.5\tabcolsep
        \xleaders#3{#2.5\@tempdimb #1{1}#2.5\@tempdimb}%
                #2\z@ plus1fil minus1fil\relax
        \hskip.5\tabcolsep}
\newcommand{\cdashlinelr}[1]{%
  \noalign{\vskip\aboverulesep
           \global\let\@dashdrawstore\adl@draw
           \global\let\adl@draw\adl@drawiv}
  \cdashline{#1}
  \noalign{\global\let\adl@draw\@dashdrawstore
           \vskip\belowrulesep}}
\makeatother

\usepackage{wrapfig}
\title{Temporal-IRL: Modeling Port Congestion and Berth Scheduling with Inverse Reinforcement Learning}

\usepackage[all]{hypcap}

\usepackage[authoryear, round]{natbib}

\usepackage{hyperref}[citecolor=magenta,linkcolor=magenta]

\usepackage{multirow, makecell, caption}\usepackage{microtype}
\usepackage{graphicx}
\usepackage{booktabs} 

\usepackage{amsmath}
\usepackage{amssymb}
\usepackage{mathtools}
\usepackage{amsthm}
\usepackage{mathrsfs}
\usepackage{nicefrac}
\usepackage{dsfont}
\usepackage{enumitem}
\usepackage{arydshln}
\usepackage{xspace}
\usepackage[capitalize,noabbrev]{cleveref}
\usepackage{subcaption}
\usepackage{wrapfig}
\usepackage{lipsum}
\usepackage{listings}
\usepackage{xcolor}
\usepackage{tcolorbox}
\usepackage{fontawesome5} 
\usepackage{amsmath}
\usepackage{amssymb}
\usepackage{mathtools}
\usepackage{amsthm}
\usepackage{bbm}
\usepackage{soul}
\tcbuselibrary{skins, breakable, listings, theorems}
\usepackage{algpseudocode}
\usepackage{setspace}

\usepackage{color}
\definecolor{lightcyan}{RGB}{210, 245, 245} 
\definecolor{lightgreen}{RGB}{220, 245, 220} 
\definecolor{deepercyan}{RGB}{130, 180, 200} 

\definecolor{lightblue}{RGB}{242, 251, 252}
\definecolor{headerbg}{RGB}{220, 230, 240} 
\definecolor{rowgray}{RGB}{245, 245, 245}
\definecolor{rowwhite}{RGB}{252, 252, 252} 
\definecolor{highlightorange}{RGB}{255, 102, 0} 

\newtcolorbox{promptbox*}[2][]{
    enhanced,              
    unbreakable,            
    before skip=2mm,        
    after skip=2mm,         
    colback=lightcyan!50!white, 
    colframe=deepercyan,    
    coltitle=white,         
    boxrule=0.5mm,          
    rounded corners,        
    arc=5pt,                
    attach boxed title to top center={yshift=-3mm}, 
    boxed title style={     
        enhanced,           
        colback=deepercyan, 
        colframe=deepercyan, 
        arc=5pt,            
        outer arc=5pt,       
        boxrule=0pt,        
    },
    title={\faBook[solid]\space #2},  
    fonttitle=\bfseries\color{white}, 
    #1,                     
    width=\dimexpr\textwidth+1\columnsep\relax, 
    left=5pt,               
    right=5pt,              
}

\newcommand\pythonstyle{\lstset{
basicstyle=\ttfamily\footnotesize,
language=Python,
morekeywords={self, clip, exp, mse_loss, uniform_sample, concatenate, logsumexp},              
keywordstyle=\color{deepblue}, 
stringstyle=\color{deepgreen},
frame=single,                         
showstringspaces=false
}}

\lstnewenvironment{python}[1][]
{
\pythonstyle
\lstset{#1}
}
{}

\newcommand\pythoninline[1]{{\pythonstyle\lstinline!#1!}}

\definecolor{promptgray}{RGB}{200,200,200}
\definecolor{promptblue}{RGB}{25,118,210}
\definecolor{darkblue}{HTML}{0C2340}
\definecolor{gold}{HTML}{AE9142}

\newtcolorbox{promptbox}[2][]{%
    enhanced,
    unbreakable,
    before skip=2mm,
    after skip=2mm,
    colback=darkblue!5!white, 
    colframe=darkblue, 
    coltitle=white, 
    boxrule=0.5mm,
    sharp corners,
    arc=5pt,
    attach boxed title to top center={yshift=-3mm},
    boxed title style={
        enhanced,
        colback=gold, 
        colframe=darkblue,
        arc=5pt,
        outer arc=5pt,
        boxrule=0pt,
    },
    title={\faLightbulb[solid]\space #2},
    fonttitle=\bfseries\color{white}, 
    #1
}

\makeatletter
\def\mathcolor#1#{\@mathcolor{#1}}
\def\@mathcolor#1#2#3{%
  \protect\leavevmode
  \begingroup
    \color#1{#2}#3%
  \endgroup
}
\makeatother

\definecolor{NDblue}{RGB}{12, 35, 64} 
\definecolor{NDgold}{RGB}{174, 145, 66} 

\hypersetup{
    colorlinks = true,    
    linkcolor = NDblue,    
    citecolor = NDgold,   
    urlcolor = NDblue,     
    filecolor = NDblue     
}

\Crefformat{equation}{#2Eq.\;(#1)#3}

\Crefformat{figure}{#2Figure #1#3}
\Crefformat{assumption}{#2Assumption #1#3}
\Crefname{assumption}{Assumption}{Assumptions}

\usepackage{crossreftools}
\usepackage{dsfont}
\usepackage{nicefrac}

\author[1,2,*]{Yikuan Hu}
\author[1,2,*]{Zixiang Xu}
\author[1,2,*]{Wei Zhang}
\author[1,2,*]{Xinyu Yang}
\author[1,2,*]{Guo Li}
\author[2]{Nikolay Aristov}
\author[1]{Mingjie Tang}
\author[2]{Elenna R Dugundji}

\affil[1]{Sichuan University}
\affil[2]{Massachusetts Institute of Technology}

\affil[*]{Equal Contribution}

\correspondingauthor{elenna\_d@mit.edu (Elenna R Dugundji)}

\begin{abstract}
\textbf{Abstract:} Predicting port congestion is crucial for maintaining reliable global supply chains. Accurate forecasts enable improved shipment planning, reduce delays and costs, and optimize inventory and distribution strategies, thereby ensuring timely deliveries and enhancing supply chain resilience. To achieve accurate predictions, analyzing vessel behavior and their stay times at specific port terminals is essential, focusing particularly on berth scheduling under various conditions. Crucially, the model must capture and learn the underlying priorities and patterns of berth scheduling. Berth scheduling and planning are influenced by a range of factors, including incoming vessel size, waiting times, and the status of vessels within the port terminal. By observing historical Automatic Identification System (AIS) positions of vessels, we reconstruct berth schedules, which are subsequently utilized to determine the reward function via Inverse Reinforcement Learning (IRL). For this purpose, we modeled a specific terminal at the Port of New York/New Jersey and developed Temporal-IRL. This Temporal-IRL model learns berth scheduling to predict vessel sequencing at the terminal and estimate vessel port stay, encompassing both waiting and berthing times, to forecast port congestion. Utilizing data from Maher Terminal spanning January 2015 to September 2023, we trained and tested the model, achieving demonstrably excellent results.
\end{abstract}

\begin{document}
\maketitle

\section{Introduction}

\subsection{Background}

As the COVID-19 pandemic ends, global trade is beginning to recover, resulting in a renewed boom in global maritime logistics \citep{strange20202020,anner2022power} . This boom poses significant challenges to port capacity and scheduling. Ports are key nodes in the U.S. supply chain network and serve as the primary entry points for goods into the country \citep{banomyong2005impact,aristov2024elucidating} . Therefore, it is hard to effectively enhance the resilience and reliability of the supply chain without an in-depth study of port logistics.

Modern AI technologies have increasingly permeated all aspects of society, from natural language understanding to complex decision-making systems, fundamentally reshaping how we solve real-world problems \citep{xu2025crosslingualpitfallsautomaticprobing,xu2025socialmazebenchmarkevaluatingsocial}. In the context of port logistics, this provides new opportunities to tackle long-standing challenges in scheduling and planning.

\subsection{Motivation}
Port congestion in the supply chain leads to additional delays and costs, reducing the reliability and efficiency of the supply chain and presenting significant challenges for transportation planning and coordination \citep{arvis2007cost} . We aim to accurately predict berth scheduling and subsequent vessel behavior to identify potential future congestion, thereby providing key information to optimize management and scheduling strategies within the supply chain. This will ensure the smooth flow of goods and enhance the resilience of the entire supply chain.

To achieve this, we need to understand the behavior and stay time of specific vessels at particular ports. Typically, obtaining berth scheduling instructions issued by the port for the upcoming period requires communicating with port personnel to understand their operational priorities. However, for a large number of ports, the manpower and time costs associated with this task are prohibitive \citep{alvarez2010methodology}. Therefore, we first need a reliable predictive algorithm to deeply understand the underlying patterns of vessel activities at ports. This algorithm should be capable of predicting berth scheduling, thereby deducing the behavior of vessels at each stage of their stay in the port and estimating the time consumed at each stage.

Berth schedules given to vessels are influenced by a complex interplay of factors, including vessel size, operator, duration of stay, and contractual agreements. While machine learning techniques have found widespread application in various transportation and logistics domains for predictive tasks \citep{heyns2019predicting,khalil2021forecasting}, directly applying traditional ML methods to berth scheduling presents significant challenges. These methods often struggle to effectively capture the intricate, and sometimes tacit, operational logic and priorities that govern port scheduling decisions. Consequently, relying on standard ML approaches can lead to suboptimal or inaccurate predictions in this context. Similarly, while Reinforcement Learning (RL) offers a powerful framework for decision-making \citep{kaelbling1996reinforcement} , designing a comprehensive and effective reward function that explicitly accounts for the multiple factors influencing the allocation of seats proves exceptionally difficult. Manually engineering such a reward function to develop a robust RL model capable of accurately reflecting real-world port operations is a substantial undertaking.

To circumvent the limitations of directly modeling complex reward structures, we innovatively employ Inverse Reinforcement Learning (IRL) \citep{ng2000algorithms}. By observing historical vessel behaviors, IRL enables us to infer the underlying reward function that implicitly guides port scheduling decisions. This approach allows us to deduce the probability distribution that best aligns with the port's operational patterns, providing a more nuanced understanding of vessel behavior and waiting times at specific ports. Based on this inferred reward function, we aim to establish a predictive model for berth scheduling, which will in turn allow us to accurately forecast vessel behavior and stay duration at each stage within the port.

\begin{figure*}[t]
  \centering
  \includegraphics[width=1.0\textwidth]{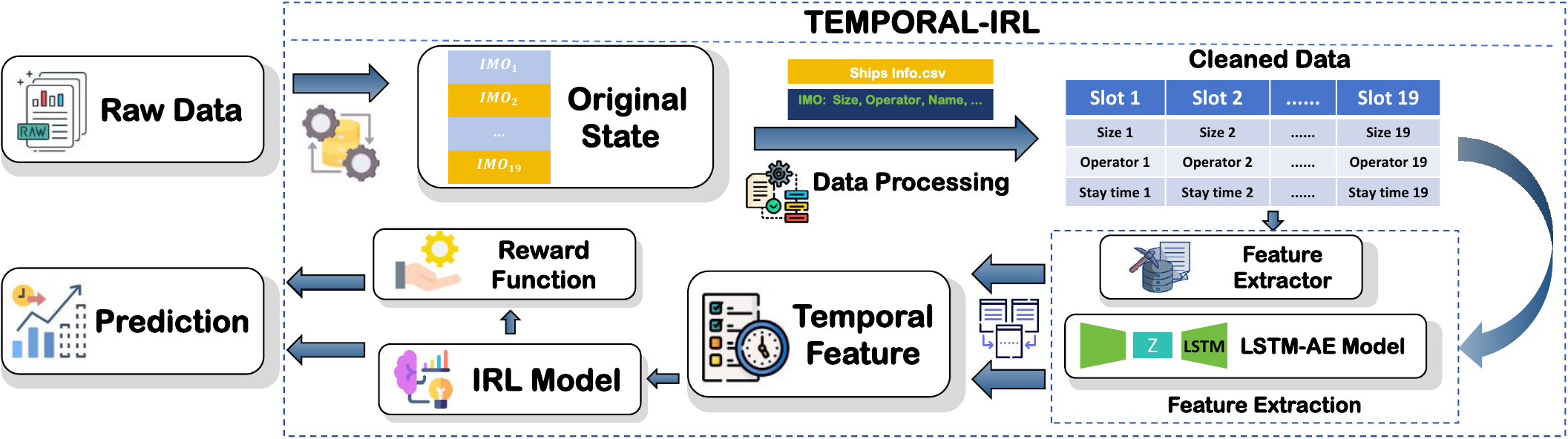}
  \caption{Overview of the proposed pipeline for port congestion forecasting using Inverse Reinforcement Learning. The pipeline processes raw data into model-ready format, extracts relevant features including temporal information, and employs an IRL model to learn a reward function for generating congestion predictions.}\label{Figure1}
\end{figure*}

\subsection{Method Summary}

Our problem can be summarized as follows: Based on the port's status and by learning expert behavior from historical data, we aim to predict the upcoming berth scheduling. This will help us determine the actions of all vessels in the model and analyze the time each vessel spends at different stages in the port. Ultimately, this will allow us to provide potential congestion forecasts.

As illustrated in \autoref{Figure1}, our work can be split into the following three components:

\begin{itemize}
\item \textbf{Data Processing}: We use AIS data with detailed information about vessels visiting the Maher Terminal from January 2015 to September 2023. Located in the Port of New York and New Jersey, Maher Terminal is one of North America's largest marine container terminals. Our dataset includes IMO numbers, arrival and departure times, vessel sizes, operating companies, and other relevant details. The AIS dataset, transmitted via radio frequencies, provides real-time positions, speeds, and other dynamic vessel information. The IMO number is a unique identifier assigned to each vessel by the International Maritime Organization for tracking.

In this phase, we focus on organizing and cleaning the data using non-machine learning methods. We removed long continuous periods (e.g., weeks) during which the port was completely empty, as they do not contribute to the accuracy of our model predictions. Additionally, we applied time windows(8 hours) to discretize the continuous time data. The cleaned dataset was then used to construct Expert Decision Trajectories, where each point represents a state-action pair that reflects the decisions made by experts in the corresponding states.

\item \textbf{IRL Feature Extraction}: We input the processed time-windowed data into the designed IRL Feature Extractor. The IRL Feature Extractor first concatenates and compresses the data from the Expert Decision Trajectories, transforming it into a format that the IRL algorithm can accept. Additionally, we have embedded an Long Short-Term Memory Recurrent Neural Networks (LSTM)\citep{hochreiter1997long} within this component to extract key temporal information from the Expert Decision Trajectories and combine it with the extracted features. The resulting features, now incorporating temporal information, are then fed into the IRL model.

\item \textbf{IRL Model}: We employed IRL algorithm to uncover the underlying instruction patterns from expert decision trajectories,  resulting in a reward function that reveals the port's instruction logic. Utilizing this reward function, we can infer the probability distribution of instructions issued to vessels under specific port states, enabling predictions of vessel actions and dwell times, and further forecasting port congestion.

\end{itemize}

Through Temporal-IRL, we can learn the patterns of berth scheduling, predict the behavior of vessels in the port, and subsequently forecast their waiting times, berth occupancy, and potential port congestion. In summary, our contributions are as follows:
\begin{itemize}
    \item We have innovatively introduced IRL algorithm to the problem of predicting berth scheduling and port congestion, addressing challenges that traditional machine learning methods often struggle to overcome.

    \item We have effectively enhanced the IRL feature extraction process by integrating temporal information, which improves the predictive capability of the IRL model for this specific problem and yields desirable results.

    \item Our proposed method can accurately predict the berth scheduling for vessels and forecast various events, such as port congestion. This provides a solid foundation for further optimization of scheduling in the upstream and downstream supply chains.
\end{itemize}
\section{State of the Practice}
\subsection{Prediction of Port Congestion}
Over the past two decades, numerous research projects have focused on predicting port congestion and addressing related traffic flow issues. These efforts have contributed potential solutions for optimizing port scheduling.

Studies have analyzed congestion at the Port of Busan by examining key metrics such as the number of waiting vessels, intervals between vessel arrivals, and berth service times \citep{yeo2007evaluation}. Another line of research has measured port congestion by investigating the relationship between the total number of container vessels at a port and the availability of container berths \citep{steven2012choosing}. Alternatively, some research has focused on calculating container flow times at U.S. ports, using extended flow times as an indicator of heightened congestion \citep{leachman2011congestion}. Despite these efforts, data for such studies often originated from port authorities and industry stakeholders, introducing potential heterogeneity and inconsistencies. Moreover, the analytical capabilities available at the time constrained these initial studies to consider only a limited range of influencing factors, thus lacking a thorough exploration of the various potential drivers of port congestion.

Beyond traditional statistical methods, innovative approaches have been integrated into transportation research. Notably, modern artificial intelligence techniques like machine learning and reinforcement learning have performed well in identifying port congestion. This section explores work conducted in these areas.

Recent approaches involve the development of an algorithm based on LSTM networks \citep{fadnes2023using,peng2023deep}, which are recurrent neural networks specifically designed for sequential and time-series data. LSTM models excel at capturing temporal dependencies and patterns over extended periods, making them well-suited for forecasting tasks. The algorithm uses multiple data types, including Automatic Identification System (AIS) data, vessel characteristics, weather data, and commodity prices, as input variables to predict future congestion levels at the Port of Paranaguá in Brazil. This method has shown excellent predictive performance and inspired us to leverage potential time series information to optimize our algorithm.

Among the diverse machine learning methods, there are approaches based on Markov chain analysis to estimate port congestion and vessel waiting times \citep{pruyn2020analysis}. Other methods apply Extreme Gradient Boosting (XGBoost) or traditional LSTM  algorithms to predict container vessels' time in port, and an Artificial Neural Network (ANN) algorithm to predict congestion risks directly related to vessel delays at port berths \citep{lamii2022using}.

However, these approaches typically treat congestion as an isolated issue for prediction. Our innovation lies in not predicting congestion directly but in fundamentally understanding the rules for issuing commands to vessels. By predicting upstream commands, we simulate future vessel movements and dwell times, ultimately providing an accurate forecast of port congestion.We will model using these methods on the same dataset in the experiment section and compare the results with our approach.

\subsection{Inverse Reinforcement Learning}
An early approach to inferring reward functions via IRL in continuous state spaces was initially presented \citep{ng2000algorithms}. Later developments incorporated the maximum entropy model \citep{ziebart2008maximum}, employing probabilistic methods to enhance the determination of reward functions and substantially improve inference performance. A reward function quantifies the preferences or objectives of an agent, providing a foundation for the learning algorithm to understand and replicate decision-making behavior.

IRL provides an effective approach to address the problems of difficult-to-determine reward functions and hard-to-capture feature patterns, which is useful for behavior prediction. Specific applications include predicting certain animal behaviors \citep{ashwood2022dynamic} and predicting human-operated vehicle behaviors \citep{higaki2022investigation}. We aim to use IRL to predict the berth schedules issued by port terminals to vessels, learning the underlying decision patterns of the ports from a large number of features, and inferring appropriate reward functions to obtain the probability distribution of the decisions, thereby achieving the desired results. According to our research, it appears that no team has yet attempted this method.

\section{Methodology}
In this section, we will outline our specific methodology. We begin by explaining how we model the entire scenario, including state composition, action design, and data processing. Next, we detail the feature extraction process and the IRL model. During feature extraction, we capture the characteristics of expert decision trajectories needed by the IRL algorithm. Using an LSTM integrated within this component, we extract feature information from time-series data, which is then fed into the IRL model. The IRL model learns the berth scheduling from historical data to infer the reward function, from which we can derive the probability distribution of decisions.

\subsection{Scene Modeling\& Data Processing}
\subsubsection{Background Information}
The Maher Terminal has six berths dedicated to unloading and loading vessels. When all berths are occupied, or if a vessel's arrival or departure needs to be coordinated with other vessels, arriving vessels must queue in a waiting zone. The entry of vessels from the waiting zone to the berths is controlled by port directives, rather than in the order of their arrival. The port considers various factors, such as the size of the vessel, the ocean carrier company, the origin of the vessel, customs information, berth size, and the depth of the channel, when making these decisions.

\begin{figure*}[t]
    \centering
       \includegraphics[scale=0.33]{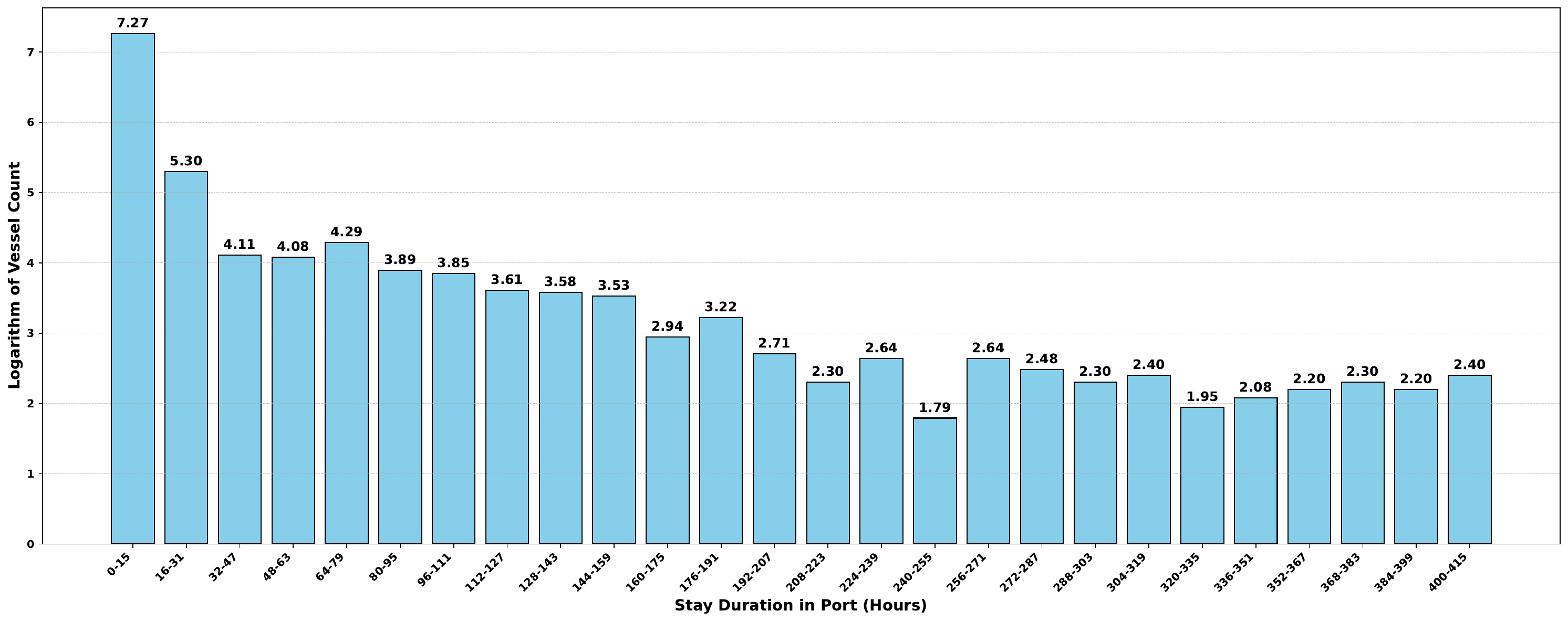}
        \caption{Port Time Proportion Analysis Chart:This chart presents the distribution of natural logarithm values for vessel counts within distinct berthing duration intervals (unit: hours). The data indicate that the majority of vessels remain in port for no more than 16 hours; therefore, the midpoint value (8 hours) was selected as the criterion for dividing the time windows.}
    \label{fig:Port_Time_Proportion_Analysis_Chart}
\end{figure*}

\subsubsection{States and Actions}
\label{sec:space definition}
To simplify analysis and improve computational feasibility, we discretized the continuous timeline into discrete time windows. We analyzed the average stay duration and selected an 8-hour interval for each time point. As shown in \autoref{fig:Port_Time_Proportion_Analysis_Chart}, the chart plots the duration of vessel stays (in hours) on the horizontal axis and the logarithmic number of vessels in each segment on the vertical axis. It visually represents the distribution of vessel stays in our dataset, showing that most vessels stay for more than 8 hours, which led us to choose an 8-hour interval for each time window.

We describe the entire port using 19 slots: the first six represent the occupancy of the port's berths, the next seven describe the occupancy of the waiting area, and the final six represent incoming vessels. From our analysis of data from January 2015 to September 2023, we found that the maximum number of vessels in the waiting area at any given time was seven, and the maximum number of incoming vessels was six. An "incoming" vessel retains this status for one time window and then either moves to the waiting area or a berth in the next time window. Based on these findings, we will proceed to model states and actions using these 19 slots.

\textbf{Definition of State Space:}
For the state of a given time window, we use a vector of length 19. Each number in this vector corresponds to the IMO number of the vessel in the respective slot; if no vessel is present, it is marked as 0. This number can later be expanded through our dataset to include more information related to the vessel. \autoref{tab:state_representation_Example} provides an example of state representation.

\begin{table}[h!]
    \caption{State Representation Example. Our state is represented as a vector of length 19, which captures the status of the port at a specific point in time.}\label{tab:state_representation_Example}
    \centering
    \begin{tabular}{l l l l}
    \textbf{slot 1} & \textbf{slot 2} & \textbf{...} & \textbf{slot 19} \\ \hline
    IMO for slot 1 (0 when empty); & IMO for slot 2 (0 when empty) & ... & IMO for slot 19  \\ \hline
    \end{tabular}
\end{table}

\begin{figure}[h!]
    \centering
    \includegraphics[scale=0.4]{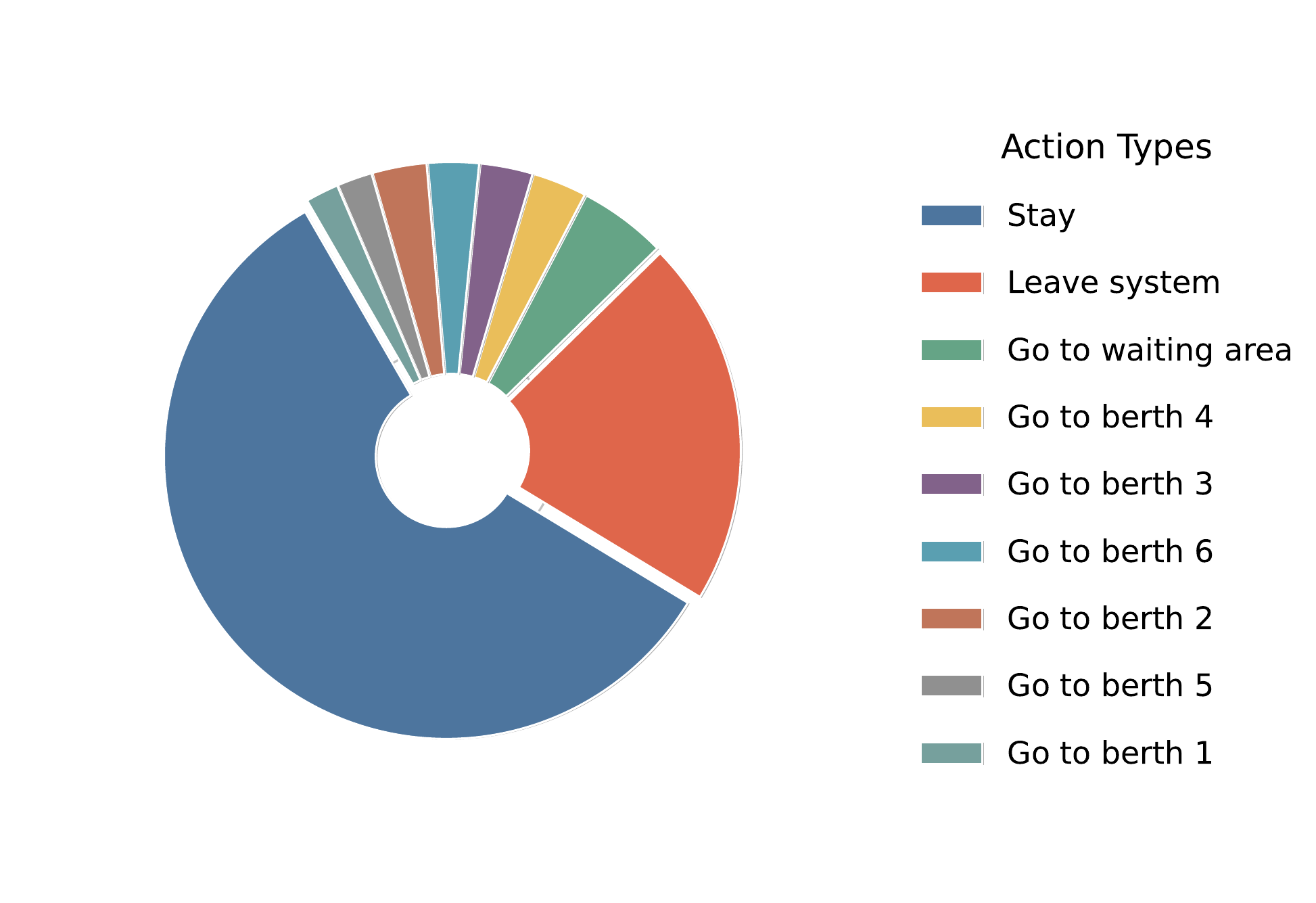}
    \caption{Distribution of nine vessel actions. Actions are displayed in descending order of frequency: 'Stay,' 'Leave system,' 'Go to waiting area,' 'Go to berth 4,' 'Go to berth 4,' 'Go to berth 3,' 'Go to berth 6,' 'Go to berth 2,' 'Go to berth 5,' and 'Go to berth 1.'}
    \label{fig:Distribution_of_Actions}
\end{figure}

\textbf{Definition of Action Space: }
For the action between two time windows, we use a vector of length 190. Each vessel has an action represented by a one-hot encoded vector of length 10. Therefore, for ten vessels, the total length of the action vector is 190. If a slot is empty, we encode it as a one-hot vector where all entries are 0, indicating "Nothing". \autoref{tab:action_representation} and \autoref{tab:action_conversion} provide examples of action representation and an action conversion table, respectively. As shown in \autoref{fig:Distribution_of_Actions}, the Distribution of Actions chart uses the names of actions as the horizontal axis and the proportion of each action's occurrence as the vertical axis. It visually displays how the quantities of actions are distributed within our action space, explaining why we chose to reduce the weights of the "Stay" action and the "Leave System" action.

\begin{table}[ht!]
    \centering
    \begin{minipage}[t]{0.48\textwidth}
        \centering
        \caption{Action Representation. our action space is also a vector of length 19, which delineates the scheduling instructions issued by the port for each vessel. The significance of the numerical values within the action space can be referenced in \autoref{tab:action_conversion}}
        \label{tab:action_representation}
        \vspace{6pt}  
        \begin{tabular}{@{}l@{}}
            \hline
            \textbf{slot 1} \\ 
            One-hot encoding of the action for slot1\\[2pt]  
            \textbf{slot 2} \\ 
            One-hot encoding of the action for slot2\\[2pt]
            \textbf{......} \\[2pt]
            \textbf{......} \\[2pt]
            \textbf{slot 19} \\ 
            One-hot encoding of the action for slot19\\ 
            \hline
        \end{tabular}
    \end{minipage}
    \hfill
    \begin{minipage}[t]{0.48\textwidth}
        \centering
        \caption{Action Conversion. This table delineates all potential scheduling behaviors for each vessel.}\label{tab:action_conversion}
        \vspace{6pt}  
        \begin{tabular}{@{}ll@{}}
            \hline
            \textbf{Index} & \textbf{Action} \\ 
            \hline
            1 & Nothing \\[2pt]
            2 & Stay \\[2pt]
            3 & Go to waiting area \\[2pt]
            4 & Go to berth1 \\[2pt]
            5 & Go to berth2 \\[2pt]
            6 & Go to berth3 \\[2pt]
            7 & Go to berth4 \\[2pt]
            8 & Go to berth5 \\[2pt]
            9 & Go to berth6 \\[2pt]
            10 & Leave system \\[2pt]
            \hline
        \end{tabular}
    \end{minipage}
\end{table}

\subsection{Feature Extraction}
To enable the model to understand more information related to the vessels, we first linked each IMO number with our dataset, expanding each single IMO entry in the state into a triplet (we will introduce this in \autoref{sec:Dataset Information}). We concatenated this triplet into a comprehensive set of information. This transformation converts the IMO number, which the model cannot inherently understand, into numerical information that represents vessel characteristics comprehensible by the model. To ensure the model can capture the temporal information present along the timeline, we input these processed features into an LSTM model capable of capturing sequential data. Following the LSTM, we added an Autoencoder\citep{bank2023autoencoders} structure to extract feature information from the LSTM's output, thus forming our LSTM-AE architecture.

\subsubsection{Obtaining Feature Information from the Dataset}
\label{sec:Dataset Information}

Since the model cannot directly process IMO information, we need to convert the IMO number into features that the model can understand before inputting the data into the LSTM-AE model. For each IMO number, we will convert it into three pieces of information by querying our dataset: 
\begin{enumerate}
    \item \textbf{Size}: The size classification of the vessel.
    \item \textbf{Carrier}: The company code of the vessel's operator, which is determined based on the ranking of companies by their market value.
    \item \textbf{Stay time}: The duration the vessel has stayed in the given slot, measured in eight-hour intervals.
\end{enumerate}

Formally, we transform the data 
\begin{center}
from 
\[
[IMO_1, IMO_2, ..., IMO_{19}]
\] 
to
\[
[size_1, company_1, staytime_1, size_2, company_2, staytime_2, ..., size_{19}, company_{19}, staytime_{19}]
\].
\end{center}
\subsubsection{Extracting Temporal Information}
\label{sec:Temporal Information}

After inputting the data processed in \autoref{sec:Dataset Information} into the model, the LSTM-AE model outputs a probability distribution for the predictions within the current time window. We calculate the weighted cross-entropy loss by comparing this output with the true action distribution. The cross-entropy loss formula is:

\begin{align}
    L = -\sum_{i=1}^{n} w_i \left( y_i \log(\hat{y}_i) + (1 - y_i) \log(1 - \hat{y}_i) \right)
\end{align}

Where \( y_i \) is the true label, \( \hat{y}_i \) is the predicted probability, \( w_i \) is the weight for each action, and \( n \) is the total number of actions. 

Based on the statistics from our cleaned data (as shown in \autoref{fig:Distribution_of_Actions}), we found that the actions "Nothing," "Stay," and "Leave System" were significantly more frequent than other actions in the action space. Therefore, we substantially reduced the weight of the "Nothing" action and partially reduced the weights of the "Stay" and "Leave System" actions (by adjusting their weights in the loss function to 0.1 and 0.3, respectively).. Then, we used the Autoencoder's hidden layer output from the trained LSTM-AE model as the extracted features, which served as new inputs for the IRL model.

\subsection{Inverse Reinforcement Learning}
IRL is a method for inferring the reward function by observing expert behavior. The fundamental concept is based on the assumption that in many real-world problems, it is challenging to directly define a reasonable reward function. However, by observing the behavior of experts (such as human operators or optimized systems), we can indirectly infer the reward function implicit in their decision-making processes. In our berth scheduling predicting scenario, expert behavior could be the historical operational records of port schedulers. Through IRL, we can learn a reward function from these expert state-action pairs that best explains their behavior, thus achieving the goal of simulating expert behavior.

The core idea of IRL is to learn a reward function \( R(s, a) \) from the expert's state-action pairs. In other words, IRL aims to find a reward function that makes the policy trained under that reward function as consistent as possible with the expert's policy. 

Maximum Entropy Inverse Reinforcement Learning (MaxEnt IRL) is a common IRL method that addresses uncertainty by introducing the principle of maximum entropy. The basic idea of the maximum entropy principle is to choose the policy with the highest entropy among all possible policies that explain the expert's behavior, thereby maximizing information and uncertainty, and avoiding making unnecessary assumptions about unknown information.In MaxEnt IRL, we assume that the expert policy \( \pi^* \) follows the following probability distribution:

\begin{align}
    \pi^*(a|s) = \frac{1}{Z(s)} \exp(Q(s, a)) 
\end{align}

This equation states that the probability of the expert choosing action \( a \) in state \( s \) is proportional to the exponential of the action-value function \( Q(s, a) \), divided by the partition function \( Z(s) \). The action-value function \( Q(s, a) \) represents the expected cumulative reward after taking action \( a \) in state \( s \). The partition function \( Z(s) \) is used to normalize the probability distribution, ensuring that the sum of the probabilities over all possible actions is 1.

\begin{align}
    Z(s) = \sum_{a} \exp(Q(s, a))
\end{align}

The partition function \( Z(s) \) is calculated by summing the exponentials of the action-value function \( Q(s, a) \) over all possible actions \( a \). This normalization ensures that the resulting probability distribution \( \pi^*(a|s) \) is valid and sums to 1.

In MaxEnt IRL, our goal is to optimize the policy by maximizing the following log-likelihood function:

\begin{align}
    \mathcal{L}(\theta) = \sum_{i=1}^N \log \pi^*(a_i|s_i) = \sum_{i=1}^N \left( Q(s_i, a_i) - \log Z(s_i) \right)
\end{align}

The log-likelihood function \( \mathcal{L}(\theta) \) measures how well the policy \( \pi^* \) explains the observed expert behavior. It is calculated by taking the logarithm of the probability distribution \( \pi^*(a|s) \) for each observed state-action pair \( (s_i, a_i) \), and summing over all \( N \) observed pairs. The term \( \log \pi^*(a_i|s_i) \) is expanded into \( Q(s_i, a_i) - \log Z(s_i) \), where \( Q(s_i, a_i) \) is the action-value function and \( \log Z(s_i) \) is the logarithm of the partition function.

To calculate \( Q(s, a) \), MaxEnt IRL uses the Bellman expectation equation:

\begin{align}
    Q(s, a) = R(s, a) + \gamma \sum_{s'} P(s'|s, a) V(s')
\end{align}

The Bellman expectation equation expresses the action-value function \( Q(s, a) \) as the immediate reward \( R(s, a) \) plus the discounted expected value of the next state \( V(s') \). Here, \( \gamma \) is the discount factor, which determines the importance of future rewards. \( P(s'|s, a) \) is the transition probability from state \( s \) to state \( s' \) given action \( a \). \( V(s') \) is the state-value function of the next state \( s' \).

The state-value function \( V(s) \) can be computed using the following equation:

\begin{align}
    V(s) = \sum_{a} \pi(a|s) Q(s, a)
\end{align}

The state-value function \( V(s) \) is the expected value of the action-value function \( Q(s, a) \), averaged over all possible actions \( a \) according to the policy \( \pi(a|s) \). This equation integrates the value of taking each action \( a \) in state \( s \), weighted by the probability of choosing that action under the policy \( \pi \).

By iteratively calculating \( Q(s, a) \) and \( V(s) \), and optimizing the log-likelihood function, we can obtain the optimal reward function \( R(s, a) \). This iterative process involves adjusting the parameters of the reward function to maximize the log-likelihood, thereby refining the reward function to accurately model the expert's behavior.

We used the state and action spaces defined in \autoref{sec:space definition} for our IRL, based on which we established expert trajectories from our data. We then used the method introduced in \autoref{sec:Dataset Information} to extract the state characteristics from the state trajectories and concatenated them with the temporal characteristics obtained in \autoref{sec:Temporal Information} and the hot encoded action characteristics. These combined features were used as the input for IRL.

Finally, we adjusted the trainable parameters of the reward function using gradient descent. This involved iteratively updating the parameters to minimize the difference between the predicted rewards and the actual rewards observed in the expert trajectories. Specifically, we calculated the gradient of the log-likelihood function with respect to the reward function parameters and used this gradient to update the parameters in the direction that maximized the likelihood of the expert's behavior. By doing so, we iteratively refined the reward function, and through these updates, we obtained a reward function that effectively modeled the expert's decision-making process.
\section{Experiment}

\subsection{Dataset}
We utilized AIS data of container vessels near Maher Terminal from 2015 to 2023, which we organized into various tables. Our data first describes the trajectory of each vessel's entry into the port, including the IMO number, vessel name, start and end times in the waiting area, entry and exit times, the vessel's position data, and the berth where the vessel stayed. We also retrieved additional information from various sources, creating tables with details such as vessel's operator and vessel size that can be looked up using the IMO number. These tables constitute our dataset.

\begin{figure}[h]
    \centering
    \includegraphics[scale=0.18]{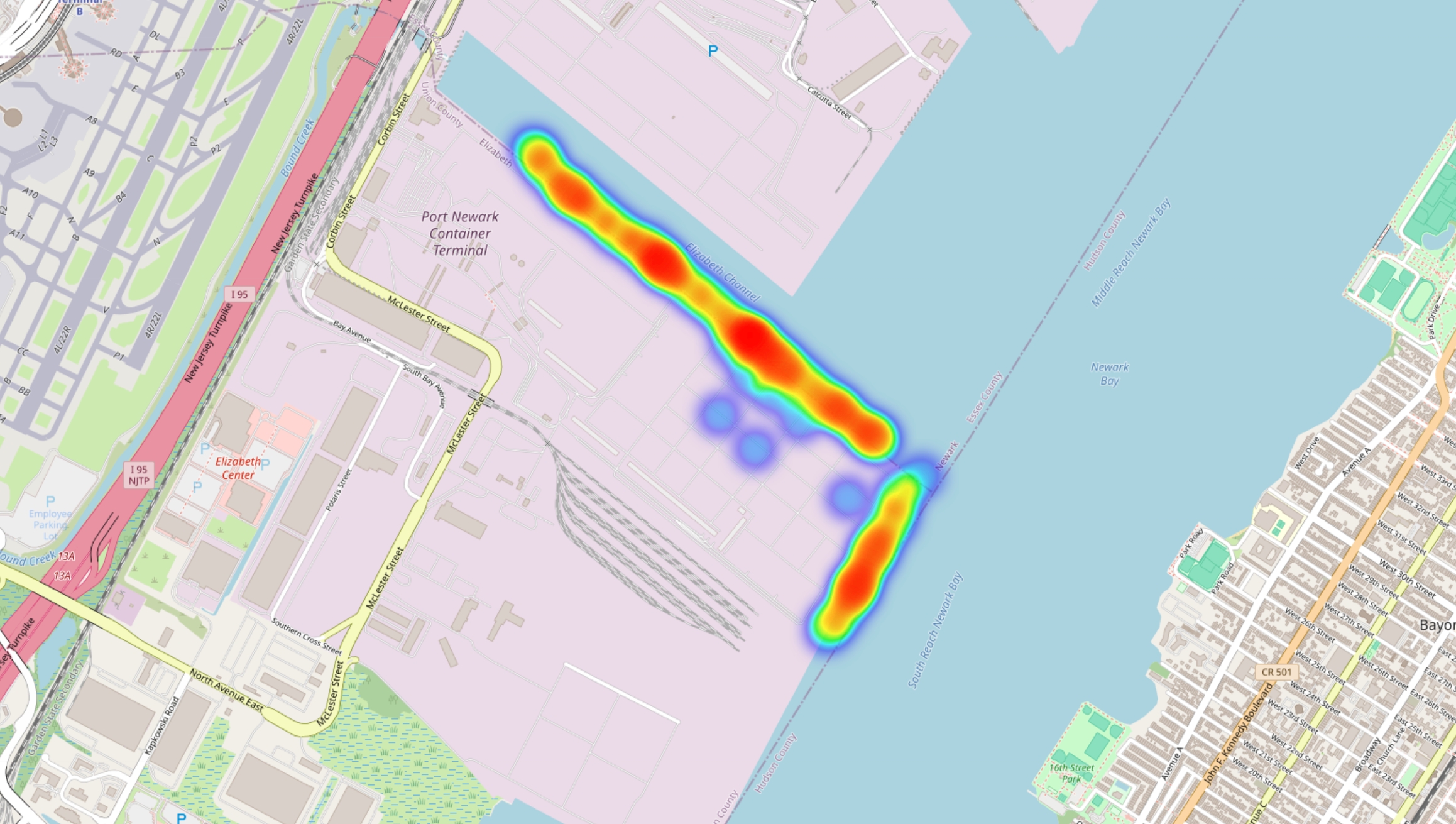}
    \caption{Heatmap of Vessel Distribution: This figure illustrates the distribution of vessels within the berths in our dataset. It is evident that the majority of vessels are dispersed across different berths, which substantiates the feasibility of constructing a streamlined and efficient modeling approach.}
    \label{fig:vesselDistribution}
\end{figure}

To provide a visualization of our dataset, we created a heatmap of ship trajectories as shown in \autoref{fig:vesselDistribution} below by analyzing the locations where vessels appeared throughout the port in our dataset. This heatmap visually represents the distribution of vessels' possible locations within the port, and in the vicinity of the port.


For the raw data, we extracted the core information that influences berth scheduling, including the vessel's IMO, operator, size, arrival time, and departure time. By defining time intervals, we structured the data to reflect temporal relationships, thus clarifying the vessel's actions at each stage.

\subsection{Results and Model Evaluation}
This section primarily presents the results of applying Temporal-IRL to the data and compares Temporal-IRL with other models.

To verify the correctness of our hypotheses regarding the use of temporal information and IRL, as well as to validate our model's functionality, we established the following testing procedure. We constructed an LSTM model without IRL, a feedforward neural network\citep{Rumelhart1986fnn} (FNN) model without temporal information, a model combining an FNN as a feature extractor with IRL and an XGBoost model. We tested the performance of these models on our dataset. Finally, we implemented an IRL model using features extracted by the LSTM-AE and applied it to the same dataset, comparing its performance with the previous models. The experimental results demonstrated that Temporal-IRL outperformed the other models, confirming our hypotheses and validating the functionality of Temporal-IRL.

\subsubsection{Data Processing for Port Behavior Prediction}
We first demonstrate the effect of feature extraction and compression using the combination of LSTM and Autoencoder (AE). When using the LSTM structure, we observed the following distribution of SHAP values, showing the contribution of each input feature to the model's output, displayed using a heatmap.

\begin{figure}[htbp]
    \centering
    \begin{subfigure}[c]{0.66\textwidth}
        \centering
        \includegraphics[width=\textwidth]{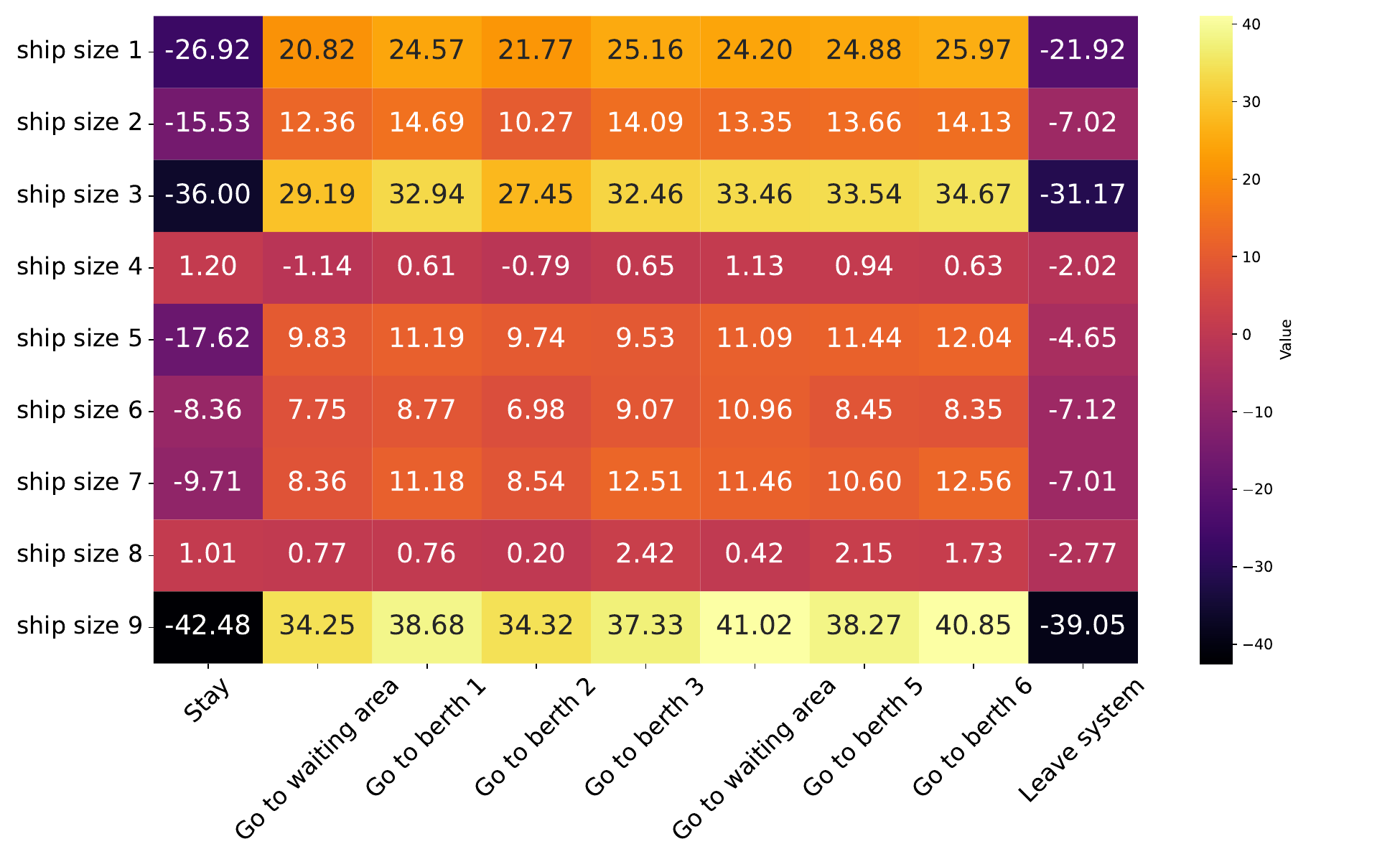}
        \caption{Size Input Feactrue SHAP Heatmap}
        \label{fig:sub1}
    \end{subfigure}
    \begin{subfigure}[c]{0.66\textwidth}
        \centering
        \includegraphics[width=\textwidth]{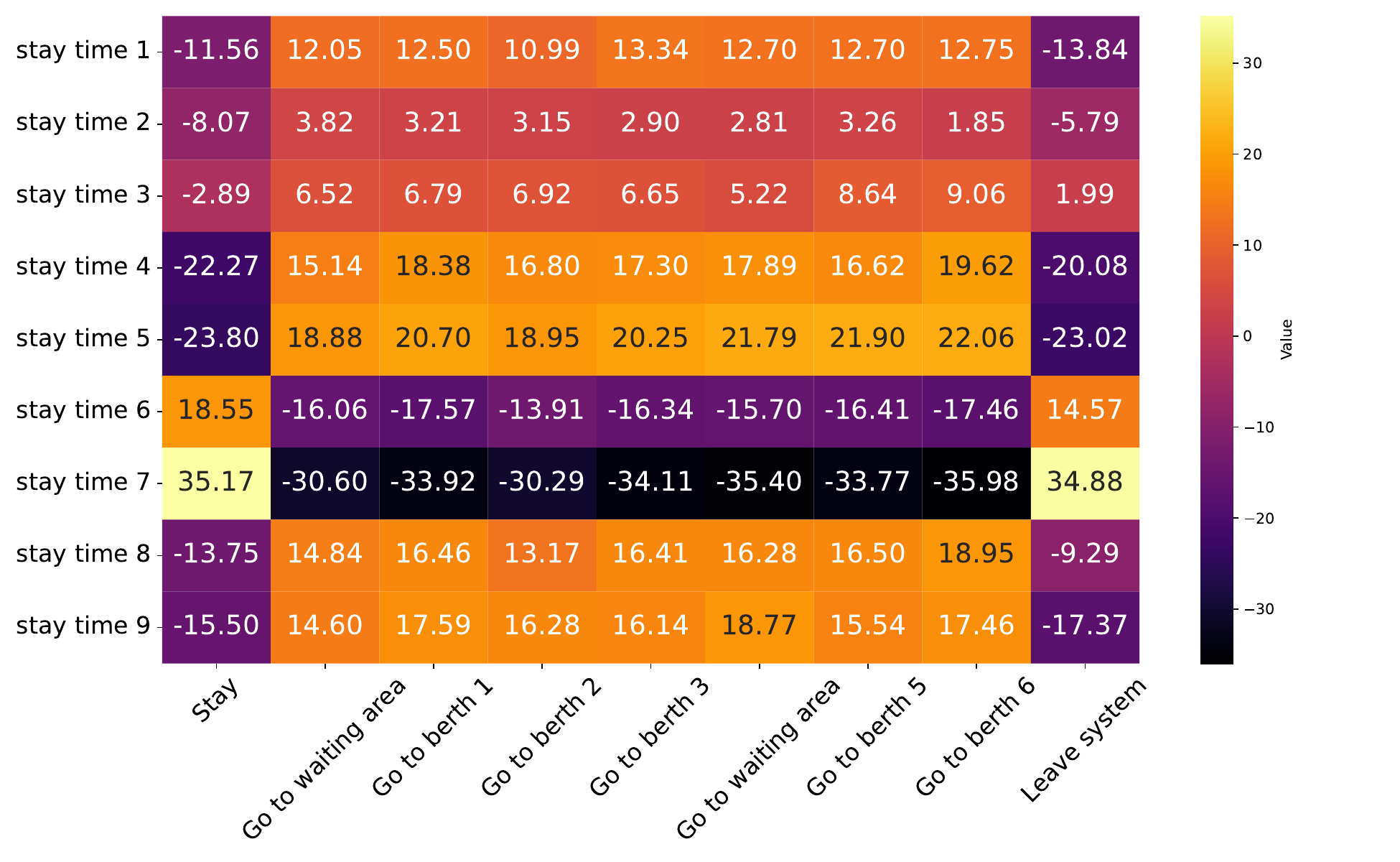}
        \caption{Stay-Time Input Feactrue SHAP Heatmap}
        \label{fig:sub2}
    \end{subfigure}
    \begin{subfigure}[c]{0.66\textwidth}
        \centering
        \includegraphics[width=\textwidth]{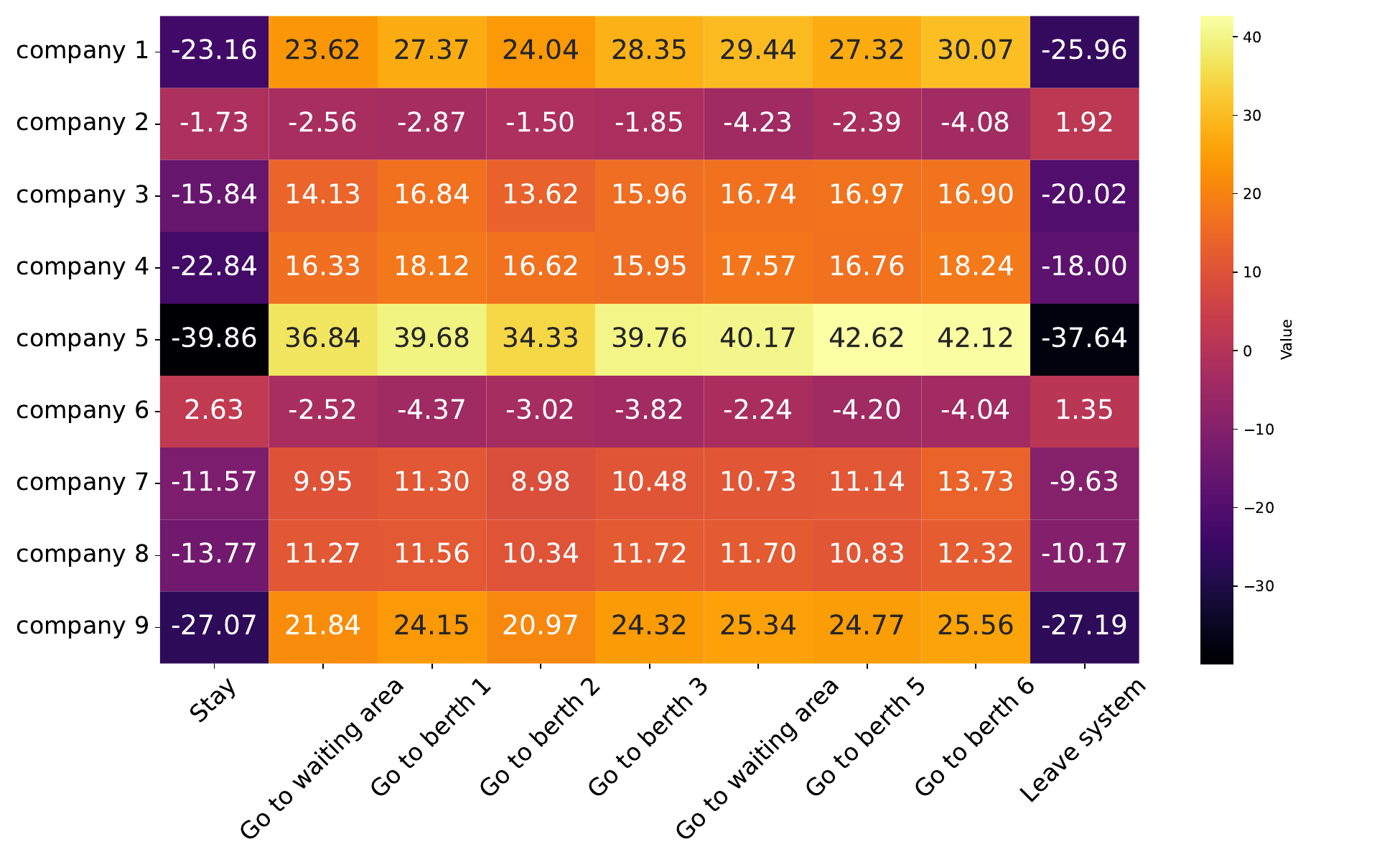}
        \caption{Company Input Feactrue SHAP Heatmap}
        \label{fig:sub3}
    \end{subfigure}
    
    \caption{The SHAP values heatmap: The SHAP values heatmap for entities 6 to 12 shows the impact of different parameters on the actions of the seven slots in the waiting area. We can observe that for different actions in different slots, the size of the ship, the company, and the stay time all have varying degrees of influence. It is difficult to directly summarize and generalize these patterns, which aligns with our judgment of the problem. For such complex and hard-to-understand underlying patterns, using IRL to learn them is an excellent strategy.}
    \label{fig:SHAP}
\end{figure}

The SHAP values heatmap \autoref{fig:SHAP} for entities 1 to 9 shows the impact of some parameters on the actions of the seven slots in the waiting area. We can observe that for different actions in different slots, the size of the vessel, the company, and the stay time all have varying degrees of influence. It is difficult to directly summarize and generalize these patterns, which aligns with our judgment of the problem. For such complex and hard-to-understand underlying patterns, using IRL to learn them is an excellent strategy.

    

\subsubsection{Prediction Results for Port Behavior}
We tested various models to predict port behavior, port congestion points, and the times of vessels entering and leaving the port. The accuracy of these predictions is calculated as follows:

\begin{equation}
\text{accuracy} = \frac{\text{Number of matching actions}}{\text{Total number of actions}}
\end{equation}

The results for these models are as follows:
\begin{table}[h!]
\caption{Action Accuracy Comparison of Different Models: The experimental results demonstrate that Temporal-IRL achieves a significant improvement in action accuracy prediction compared to other baseline models.}\label{tab:versions}
\centering
\begin{tabular}{l l l}
models & action accuracy \\
\hline

LSTM & 70.13\% \\
FNN-IRL & 70.01\% \\
FNN & 55.83\% \\        
\hline
Temporal-IRL & \textbf{82.64}\% \\
\hline
\end{tabular}

\label{table
}
\end{table}

\subsubsection{Prediction Results for Port Events}
We tested various models for predicting port congestion and the times of vessels leaving the port. The accuracy of these predictions is calculated as follows:

\begin{equation}
\text{accuracy} = \frac{\text{Number of correct predictions}}{\text{Total number of predictions}}
\\
\end{equation}

\begin{table}[t]
    \caption{Congestion and Leave Accuracy Comparison of Different Models: The experimental results demonstrate that Temporal-IRL achieves state-of-the-art performance in predicting leave accuracy while approaching comparable state-of-the-art levels in congestion accuracy prediction.}\label{tab:versions}
    \begin{center}
    \begin{tabular}{l l l l}
        models & congestion accuracy & leave accuracy \\
        \hline
        LSTM & 62.85\% & 61.07\% \\
        FNN-IRL & 61.31\% & 60.20\% \\
        FNN & 57.62\% & 39.58\% \\
        XGBoost & \textbf{75.89}\% & 66.01\% \\
        \hline
        Temporal-IRL & 74.06\% & \textbf{71.51}\% \\
        \hline
    \end{tabular}
    \end{center}
    
\end{table}
\begin{wrapfigure}{r}{0.5\textwidth}
    \centering
    \vspace{-2em}
    \includegraphics[width=\linewidth]{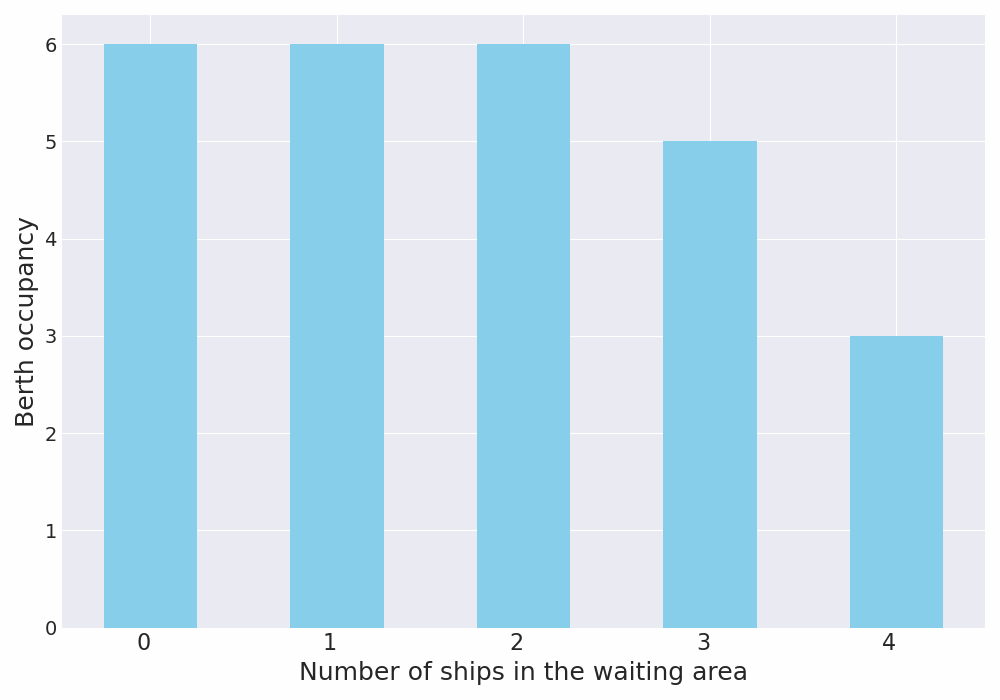}
    \caption{The Congestion Statistic: This figure illustrates the relationship between the number of vessels in the waiting area and berth occupancy status at the port. When the number of vessels in the waiting area exceeds three, the berths are not fully occupied. This phenomenon may be attributed to congestion caused by an excessive number of vessels in the waiting area, which could potentially lead to operational inefficiencies in berth allocation and scheduling processes.}
    \label{fig:conjecture}
    \vspace{-3em}
\end{wrapfigure}

As shown in \autoref{fig:conjecture}, through analyzing the data, we observed that when the number of vessels in the waiting area is greater than or equal to three, there are still available berths.  In contrast, when the number of vessels in the waiting area is fewer than two, no available berths are present.One potential reason for the observed phenomenon where vessels remain in the waiting area despite the availability of berths is that congestion in the access channels from the waiting area to the berths impedes the port's ability to issue berth assignment instructions to the vessels.

Based on this observation,we define congestion as the presence of three or more vessels in the waiting area at the same time.In each prediction, the model's output actions are applied to the current state to forecast the next congestion state. The predicted congestion state is then compared to the actual congestion state. If the states match, the prediction is considered correct; otherwise, it is deemed incorrect.

Similarly, we define the departure time window as the time window when a vessel leaves the port. We compare the actual departure events of the vessel with the predicted time window for each departure. If they match, it is considered a correct prediction; otherwise, it is deemed incorrect.

\autoref{tab:versions} presents a comparative analysis of our Temporal-IRL method against other models for both congestion and leave accuracy prediction tasks. As clearly demonstrated, our Temporal-IRL approach exhibits a significant advantage over simpler methods like LSTM and FNN-IRL(a network using FNN to replace our LSTM feature extraction module) in predicting both port congestion and vessel departure events. This improvement underscores the effectiveness of Inverse Reinforcement Learning in capturing and leveraging temporal dependencies inherent in port operational data to achieve more accurate forecasts. Notably, while Temporal-IRL achieves a congestion prediction accuracy of 74.06\%, which is marginally lower than the XGBoost model's leading 75.89\%, it substantially surpasses both LSTM (62.85\%) and FNN-IRL (61.31\%) in this metric. This indicates that IRL-based methods, particularly Temporal-IRL, are capable of effectively discerning patterns relevant to congestion prediction. Furthermore, in the critical task of predicting vessel departure events (leave accuracy), Temporal-IRL achieves a state-of-the-art accuracy of \textbf{71.51\%}. This figure demonstrably outperforms all other models, including XGBoost (66.01\%), LSTM (61.07\%), and FNN-IRL (60.20\%), establishing Temporal-IRL as the most accurate method for forecasting vessel departures among those evaluated.
\section{Conclusion}

This study views terminal prediction as an inverse problem based on temporal information. Instead of directly predicting future events, the research focuses on inferring hidden rules or patterns from existing temporal information, such as historical data. This process, known as an 'inverse problem' involves deducing the underlying factors or rules (like terminal operation instructions) from observed results (such as vessel behavior and timing information). To enhance the predictive capability of the model, we innovatively introduced a neural network that combines LSTM and Autoencoder to focus on temporal information. This approach extracts temporal features through the neural network and then uses this information for inverse reinforcement learning to predict and infer terminal behaviors. The LSTM structure is used to extract temporal data, while IRL, compared to traditional feedforward neural networks, better mimics behavior and understands the reward structure behind observed actions, rather than merely predicting future states.

We partitioned the data from Maher Terminal, spanning from January 2015 to September 2023, according to its temporal sequence into training and testing sets and subsequently training the model on the training set and evaluating it on the testing set.We designed various evaluation metrics, such as the accuracy of predicting terminal instructions and the correctness of predicting terminal congestion states. The results indicate that Temporal-IRL demonstrates strong performance in predicting the behavior of individual terminals.

With Temporal-IRL, we can predict and model the behavior of a terminal, using our predictions to forecast various possible future scenarios for the terminal. These predictive data not only help us quickly understand terminal behavior but can also assist in optimizing terminal scheduling and modeling global supply chains in the future.

However, our study has some limitations that should be addressed. For instance, the applicability of our data is somewhat narrow, as the model has been tested on only one terminal. Applying the model framework to other terminals or scenarios could lead to broader and more practical insights. Additionally, the initial features used in Temporal-IRL are somewhat limited; incorporating more data, such as weather conditions, could potentially enhance the model's performance. Moreover, our model currently predicts based on discretized time windows, which makes it difficult to accurately capture complex behaviors within a single window. We are exploring the possibility of refining the time windows or adjusting the model to better capture the intricate behaviors within each window. Future research should expand the scope of prediction and modeling, for example, by modeling throughput behaviors across all terminals in a country or region. We could also explore alternative methods for port scheduling to better capture complex behaviors that occur within a time window.
\section{Acknowledgements}
We would like to thank the Massachusetts Institute of Technology Global Supply Chain and Logistics Excellence (SCALE) Network at the MIT Center for Transportation and Logistics for giving us the opportunity to conduct this research as part of the MIT AI Summer School for the Ningbo China Institute for Supply Chain Innovation (NISCI). Elenna Dugundji and Nikolay Aristov would also like to gratefully acknowledge discussion with Dr. Tom Vu in shaping the study conception.

\bibliography{reference}

\begin{thebibliography}{25}
\providecommand{\natexlab}[1]{#1}
\providecommand{\url}[1]{\texttt{#1}}
\expandafter\ifx\csname urlstyle\endcsname\relax
  \providecommand{\doi}[1]{doi: #1}\else
  \providecommand{\doi}{doi: \begingroup \urlstyle{rm}\Url}\fi

\bibitem[Alvarez et~al.(2010)Alvarez, Longva, and Engebrethsen]{alvarez2010methodology}
J~Fernando Alvarez, Tore Longva, and Erna~S Engebrethsen.
\newblock A methodology to assess vessel berthing and speed optimization policies.
\newblock \emph{Maritime economics \& logistics}, 12:\penalty0 327--346, 2010.

\bibitem[Anner(2022)]{anner2022power}
Mark Anner.
\newblock Power relations in global supply chains and the unequal distribution of costs during crises: Abandoning garment suppliers and workers during the covid-19 pandemic.
\newblock \emph{International Labour Review}, 161\penalty0 (1):\penalty0 59--82, 2022.

\bibitem[Aristov et~al.(2024)Aristov, Li, Koch, and Dugundji]{aristov2024elucidating}
Nikolay Aristov, Ziyan Li, Thomas Koch, and Elenna~R Dugundji.
\newblock Elucidating us import supply chain dynamics.
\newblock \emph{Procedia Computer Science}, 238:\penalty0 216--223, 2024.

\bibitem[Arvis et~al.(2007)Arvis, Raballand, and Marteau]{arvis2007cost}
Jean-Fran{\c{c}}ois Arvis, Gael Raballand, and Jean-Fran{\c{c}}ois Marteau.
\newblock \emph{The cost of being landlocked: logistics costs and supply chain reliability}, volume 4258.
\newblock World Bank Publications, 2007.

\bibitem[Ashwood et~al.(2022)Ashwood, Jha, and Pillow]{ashwood2022dynamic}
Zoe Ashwood, Aditi Jha, and Jonathan~W Pillow.
\newblock Dynamic inverse reinforcement learning for characterizing animal behavior.
\newblock \emph{Advances in neural information processing systems}, 35:\penalty0 29663--29676, 2022.

\bibitem[Bank et~al.(2023)Bank, Koenigstein, and Giryes]{bank2023autoencoders}
Dor Bank, Noam Koenigstein, and Raja Giryes.
\newblock Autoencoders.
\newblock \emph{Machine learning for data science handbook: data mining and knowledge discovery handbook}, pages 353--374, 2023.

\bibitem[Banomyong*(2005)]{banomyong2005impact}
Ruth Banomyong*.
\newblock The impact of port and trade security initiatives on maritime supply-chain management.
\newblock \emph{Maritime Policy \& Management}, 32\penalty0 (1):\penalty0 3--13, 2005.

\bibitem[Fadnes and Harviken(2023)]{fadnes2023using}
Eirik Fadnes and Espen~{\AA}sheim Harviken.
\newblock Using machine learning to predict port congestion: A study of the port of paranagu{\'a}.
\newblock Master's thesis, 2023.

\bibitem[Heyns et~al.(2019)Heyns, Uniyal, Dugundji, Tillema, and Huijboom]{heyns2019predicting}
Emiliano Heyns, S~Uniyal, Elenna Dugundji, F~Tillema, and C~Huijboom.
\newblock Predicting traffic phases from car sensor data using machine learning.
\newblock \emph{Procedia computer science}, 151:\penalty0 92--99, 2019.

\bibitem[Higaki et~al.(2022)Higaki, Hashimoto, and Yoshioka]{higaki2022investigation}
Takefumi Higaki, Hirotada Hashimoto, and Hitoshi Yoshioka.
\newblock Investigation and imitation of human captains' maneuver using inverse reinforcement learning.
\newblock 36:\penalty0 137--148, 2022.

\bibitem[Hochreiter and Schmidhuber(1997)]{hochreiter1997long}
Sepp Hochreiter and J{\"u}rgen Schmidhuber.
\newblock Long short-term memory.
\newblock \emph{Neural computation}, 9\penalty0 (8):\penalty0 1735--1780, 1997.

\bibitem[Kaelbling et~al.(1996)Kaelbling, Littman, and Moore]{kaelbling1996reinforcement}
Leslie~Pack Kaelbling, Michael~L Littman, and Andrew~W Moore.
\newblock Reinforcement learning: A survey.
\newblock \emph{Journal of artificial intelligence research}, 4:\penalty0 237--285, 1996.

\bibitem[Khalil et~al.(2021)Khalil, Amrit, Koch, and Dugundji]{khalil2021forecasting}
Sergey Khalil, Chintan Amrit, Thomas Koch, and Elenna Dugundji.
\newblock Forecasting public transport ridership: Management of information systems using cnn and lstm architectures.
\newblock \emph{Procedia Computer Science}, 184:\penalty0 283--290, 2021.

\bibitem[LAMII et~al.(2022)LAMII, FRI, MABROUKI, et~al.]{lamii2022using}
NABIL LAMII, MOUHSENE FRI, CHARIF MABROUKI, et~al.
\newblock Using artificial neural network model for berth congestion risk prediction.
\newblock \emph{IFAC-PapersOnLine}, 55\penalty0 (12):\penalty0 592--597, 2022.

\bibitem[Leachman and Jula(2011)]{leachman2011congestion}
Robert~C Leachman and Payman Jula.
\newblock Congestion analysis of waterborne, containerized imports from asia to the united states.
\newblock \emph{Transportation Research Part E: Logistics and Transportation Review}, 47\penalty0 (6):\penalty0 992--1004, 2011.

\bibitem[Ng et~al.(2000)Ng, Russell, et~al.]{ng2000algorithms}
Andrew~Y Ng, Stuart Russell, et~al.
\newblock Algorithms for inverse reinforcement learning.
\newblock In \emph{Icml}, volume~1, page~2, 2000.

\bibitem[Peng et~al.(2023)Peng, Bai, Yang, Yuen, and Wu]{peng2023deep}
Wenhao Peng, Xiwen Bai, Dong Yang, Kum~Fai Yuen, and Junfeng Wu.
\newblock A deep learning approach for port congestion estimation and prediction.
\newblock \emph{Maritime Policy \& Management}, 50\penalty0 (7):\penalty0 835--860, 2023.

\bibitem[Pruyn et~al.(2020)Pruyn, Kana, and Groeneveld]{pruyn2020analysis}
JFJ Pruyn, AA~Kana, and WM~Groeneveld.
\newblock Analysis of port waiting time due to congestion by applying markov chain analysis.
\newblock In \emph{Maritime Supply Chains}, pages 69--94. Elsevier, 2020.

\bibitem[Rumelhart et~al.(1986)Rumelhart, Hinton, and Williams]{Rumelhart1986fnn}
David~E. Rumelhart, Geoffrey~E. Hinton, and Ronald~J. Williams.
\newblock Learning representations by back-propagating errors.
\newblock \emph{Nature}, 323\penalty0 (6088):\penalty0 533--536, 1986.

\bibitem[Steven and Corsi(2012)]{steven2012choosing}
Adams~B Steven and Thomas~M Corsi.
\newblock Choosing a port: An analysis of containerized imports into the us.
\newblock \emph{Transportation Research Part E: Logistics and Transportation Review}, 48\penalty0 (4):\penalty0 881--895, 2012.

\bibitem[Strange(2020)]{strange20202020}
Roger Strange.
\newblock The 2020 covid-19 pandemic and global value chains.
\newblock \emph{Journal of Industrial and Business Economics}, 47\penalty0 (3):\penalty0 455--465, 2020.

\bibitem[Xu et~al.(2025{\natexlab{a}})Xu, Wang, Huang, Chen, Zhao, Jiang, and Zhang]{xu2025crosslingualpitfallsautomaticprobing}
Zixiang Xu, Yanbo Wang, Yue Huang, Xiuying Chen, Jieyu Zhao, Meng Jiang, and Xiangliang Zhang.
\newblock Cross-lingual pitfalls: Automatic probing cross-lingual weakness of multilingual large language models, 2025{\natexlab{a}}.
\newblock URL \url{https://arxiv.org/abs/2505.18673}.

\bibitem[Xu et~al.(2025{\natexlab{b}})Xu, Wang, Huang, Ye, Zhuang, Song, Gao, Wang, Chen, Zhou, Li, Pan, Zhao, Zhao, Zhang, and Chen]{xu2025socialmazebenchmarkevaluatingsocial}
Zixiang Xu, Yanbo Wang, Yue Huang, Jiayi Ye, Haomin Zhuang, Zirui Song, Lang Gao, Chenxi Wang, Zhaorun Chen, Yujun Zhou, Sixian Li, Wang Pan, Yue Zhao, Jieyu Zhao, Xiangliang Zhang, and Xiuying Chen.
\newblock Socialmaze: A benchmark for evaluating social reasoning in large language models, 2025{\natexlab{b}}.
\newblock URL \url{https://arxiv.org/abs/2505.23713}.

\bibitem[Yeo et~al.(2007)Yeo, Roe, and Soak]{yeo2007evaluation}
Gi-Tae Yeo, Michael Roe, and Sang-Moon Soak.
\newblock Evaluation of the marine traffic congestion of north harbor in busan port.
\newblock \emph{Journal of waterway, port, coastal, and ocean engineering}, 133\penalty0 (2):\penalty0 87--93, 2007.

\bibitem[Ziebart et~al.(2008)Ziebart, Maas, Bagnell, Dey, et~al.]{ziebart2008maximum}
Brian~D Ziebart, Andrew~L Maas, J~Andrew Bagnell, Anind~K Dey, et~al.
\newblock Maximum entropy inverse reinforcement learning.
\newblock In \emph{Aaai}, volume~8, pages 1433--1438. Chicago, IL, USA, 2008.

\end{thebibliography}

\newpage
\appendix
\onecolumn

\end{document}